\documentclass[journal]{IEEEtran}
\usepackage{mathtools}
\usepackage{amssymb}

\usepackage{dsfont}

\usepackage{cite}
\usepackage{amsmath,amssymb,amsfonts}
\usepackage{amsmath}
\newcommand{\R}{\mathbb{R}}
\usepackage{mathrsfs}
\usepackage{graphicx}
\usepackage{textcomp}
\usepackage[table,xcdraw]{xcolor}
\usepackage{ragged2e}
\usepackage{bbm}
\usepackage{multirow}
\usepackage[linesnumbered, ruled]{algorithm2e}
\usepackage{algpseudocode}
\usepackage[caption=false, font=footnotesize]{subfig}
\usepackage{mathtools} 
\usepackage[hang,flushmargin]{footmisc}
\usepackage{lipsum}
\makeatletter
\newcommand{\algorithmfootnote}[2][\footnotesize]{%
  \let\old@algocf@finish\@algocf@finish
  \def\@algocf@finish{\old@algocf@finish
    \leavevmode\rlap{\begin{minipage}{\linewidth}
    #1#2
    \end{minipage}}%
  }%
}
\usepackage[flushleft]{threeparttable}
\usepackage{etoolbox}
\appto\TPTnoteSettings{\footnotesize}

\usepackage{hyperref}
\hypersetup{
    colorlinks=true,
    linkcolor=blue,
    filecolor=magenta,      
    urlcolor=cyan,
}

\newcommand{\thickhline}{%
    \noalign {\ifnum 0=`}\fi \hrule height 1pt
    \futurelet \reserved@a \@xhline
}
\newcolumntype{"}{@{\hskip\tabcolsep\vrule width 1pt\hskip\tabcolsep}}

\usepackage{tabu}

\makeatother
\usepackage[british]{babel}
\usepackage[autostyle]{csquotes}
\usepackage{placeins}

\usepackage{float}

\def\BibTeX{{\rm B\kern-.05em{\sc i\kern-.025em b}\kern-.08em
    T\kern-.1667em\lower.7ex\hbox{E}\kern-.125emX}}

\begin{document}
\title{Automatic Diagnosis of Malaria from Thin Blood Smear Images using Deep Convolutional Neural Network with Multi-Resolution Feature Fusion}

\author{Tanvir~Mahmud,~\IEEEmembership{Student~Member,~IEEE,}
        and~Shaikh~Anowarul~Fattah,~\IEEEmembership{Senior~Member,~IEEE}
        
\thanks{T.~Mahmud, and S.~A.~Fattah are with the Department of Electrical and Electronic Engineering, BUET, Dhaka 1000,
Bangladesh (e-mail: tanvirmahmud@eee.buet.ac.bd; fattah@eee.buet.ac.bd).}
\thanks{Manuscript received March X, 2020; revised XX, 2020.}}

\maketitle

\begin{abstract}
Malaria, a life-threatening disease, infects millions of people every year throughout the world demanding faster diagnosis for proper treatment before any damages occur. In this paper, an end-to-end deep learning-based approach is proposed for faster diagnosis of malaria from thin blood smear images by making efficient optimizations of features extracted from diversified receptive fields. Firstly, an efficient, highly scalable deep neural network, named as DilationNet, is proposed that incorporates features from a large spectrum by varying dilation rates of convolutions to extract features from different receptive areas. Next, the raw images are resampled to various resolutions to introduce variations in the receptive fields that are used for independently optimizing different forms of DilationNet scaled for different resolutions of images. Afterward, a feature fusion scheme is introduced with the proposed DeepFusionNet architecture for jointly optimizing the feature space of these individually trained networks operating on different levels of observations. All the convolutional layers of various forms of DilationNets that are optimized to extract spatial features from different resolutions of images are directly transferred to provide a variegated feature space from any image. Later, joint optimization of these spatial features is carried out in the DeepFusionNet to extract the most relevant representation of the sample image. This scheme offers the opportunity to explore the feature space extensively by varying the observation level to accurately diagnose the abnormality. Intense experimentations on a publicly available dataset show outstanding performance with accuracy over 99.5\% outperforming other state-of-the-art approaches.

\end{abstract}


\section{Introduction}
\IEEEPARstart{M}{alaria}, an infectious severe blood disease transmitted through the bites of infected female \emph{Anopheles} mosquitoes, has become a potential threat for almost half of the world's population~\cite{w3}. According to WHO, in $2018$, almost $228$ million cases of malaria were estimated with $405,000$ deaths throughout the world. If not treated withing $24$ hours, some forms of malaria may cause irreversible damages, often lead to death~\cite{w1}. This makes the early diagnosis of malaria very crucial for proper treatment~\cite{w2}. Manual inspection of thick blood smear images by the experts is the widely used approach for malaria diagnosis~\cite{test}. This method heavily depends on the human expertise, is prone to error and quite burdensome as in most cases experts are unavailable in rural, under-developed areas where the disease becomes more threatening~\cite{w4}. For early diagnosis of this disease, an automated diagnosis scheme can be an effective solution to reduce the clinical burden.

\begin{figure*}[!t]
\centering
\includegraphics[scale= 0.58]{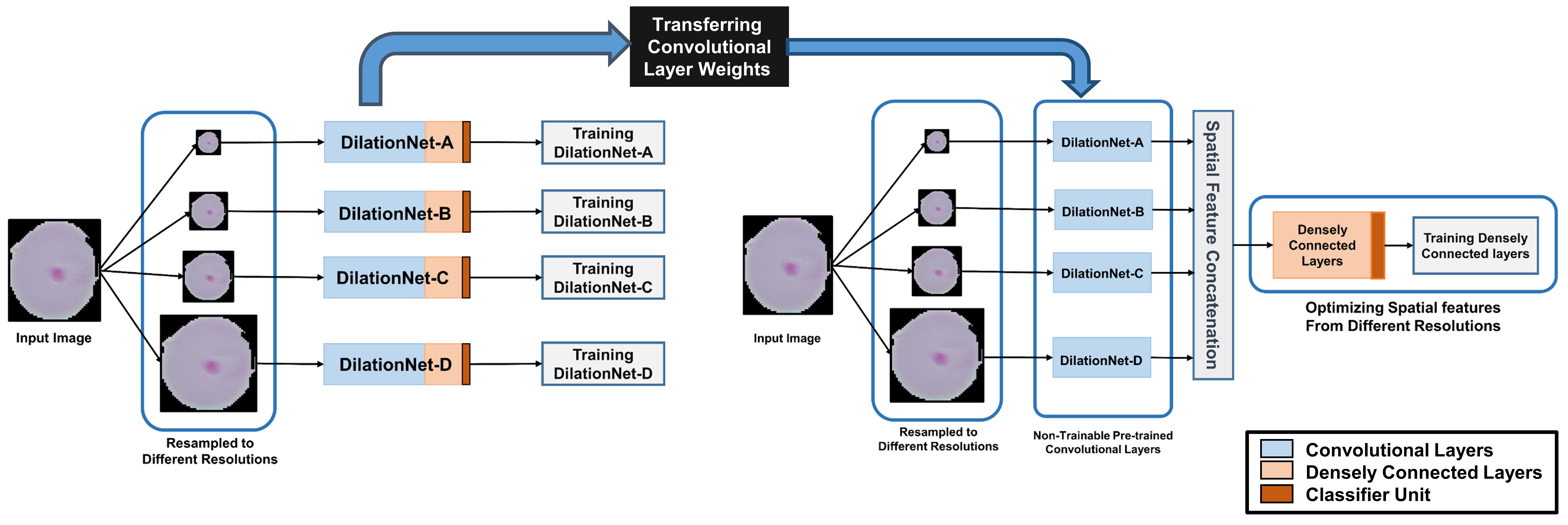}
\caption{\textbf{Schematic representation of the proposed scheme: Different form of DilationNets are trained with various resolutions of images. The initially trained convolutional layers are directly transferred for joint optimization with DeepFusionNet. Here,  DilationNet-A is optimized for resolutions $(32\times32)$, DilationNet-B for $(64\times64)$, DilationNet-C for $(128\times128)$, and DilationNet-D for $(256\times256)$.}}
\label{f1}
\end{figure*}

Large varieties of approaches have been explored for diagnosing malaria automatically ranging from traditional handcrafted feature-based methods to data-driven end-to-end deep learning approaches. In~\cite{f1,f2,f3,f4}, SVM, naive bias classifier, K-nearest neighbour and some other morphological approaches are proposed for detecting malaria parasites. In~\cite{s1,s2,s3}, some other statistical approaches are introduced for feature extraction. However, these approaches heavily depend on the manual selection and optimization of features that make the process very sensitive, often lead to poor diagnosis in the real scenario. Moreover, these methods also demand significant domain expertise that makes the process complicated for practical purposes. In~\cite{belief}, a deep belief network is introduced for the optimization of the manually extracted features. Nevertheless, this process also demands a complicated feature extraction process. Recently, several end-to-end deep learning techniques are proposed to make this feature extraction process automated and thus to make the process robust. In~\cite{dong}, some traditional convolutional neural networks, initially developed for other computer vision tasks, are utilized for malaria diagnosis. In~\cite{nih, c2, c3, c4}, different customized architectures are introduced for proper optimization. In~\cite{c2}, the transfer learning concept is employed for utilizing the VGG-$19$ architecture initially trained on ImageNet dataset~\cite{imagenet}, along with the SVM classifier. However, this results in a very large model that was optimized for a huge number of training images targeting other applications and thus becomes burdensome for mobile applications.

In this paper, an automated malaria diagnosis scheme is proposed from thin blood smear images using deep neural network. The proposed scheme is mainly focused to optimize detailed, localized features extracted from a smaller receptive field along with the generalized features extracted from broader receptive area to encompass a large spectrum. Firstly, the sample images are resampled to various resolutions to manually introduce the variations of information in similar receptive fields. Afterward, a scalable network, named as DilationNet, is introduced that can be effectively optimized for different resolutions of images. In the DilationNet, the receptive area is further varied in series of convolution operations with varying dilation rates that efficiently utilizes the filtering operation to extract features from a broader spectrum. After individually optimizing these forms of DilationNet with multi-resolution images,  a joint optimization technique is introduced for further exploration of the extracted features. As all the initial convolutional filters are efficiently optimized for different resolutions of images, these pre-trained layers are directly integrated to provide a variegated feature space incorporating different levels of observations. Later, another deep network, named as DeepFusionNet, is introduced that utilizes some additional densely connected layers to jointly optimize these extracted spatial features from different DilationNets. Further exploration of features through the joint training with the DeepFusionNet provides the opportunity to accurately diagnose the smaller, localized abnormalities as well as the scattered abnormalities spreading over a larger region. A publicly available dataset is used for extensive experimentations to validate the proposed schemes that offer state-of-the-art performance outperforming other established methods.    

\section{Methodology}
The proposed scheme trains deep neural networks in two separate stages. At the first stage, individual networks are optimized to operate with different resolutions of images. Later, a feature fusion scheme is introduced to jointly optimize all these networks through a fusion network in the final training stage. The complete scheme is shown in Fig.~\ref{f1}. All the individual portions will be discussed in detail.

\subsection{Preprocessing}
Thin blood smear images undergoes through minimal preprocessing before feeding the data in the neural network to make the process more flexible and faster in real-time applications while reducing extra computational complexity. Let consider the dataset as, $D= \{(X_i, Y_i)|\ i=1, 2, \dots, N \}$. Here, $X_i$ and $Y_i$ represent the $i_{th}$ sample image with dimension of $(h_i\times w_i)$ and corresponding level, respectively. Hence, $X_i \in \R^{h_i\times w_i}$, and $Y_i \in \{0,1\}$. These set of sample images, $\mathbf{X}$, are resampled to different resolutions, $\mathbf{X'}$, such that $X'_i \in \R^{r\times r}$, where $r \in \{32, 64, 128, 256 \}$ represents the resolution of the resampled images. Later, all the resampled images are normalized. These sets of resampled images, $\mathbf{X'}$, are used for training and evaluation. 

\subsection{Proposed DilationNet architecture for individual training:}
\begin{figure}[!t]
\centering
\includegraphics[scale= 0.6]{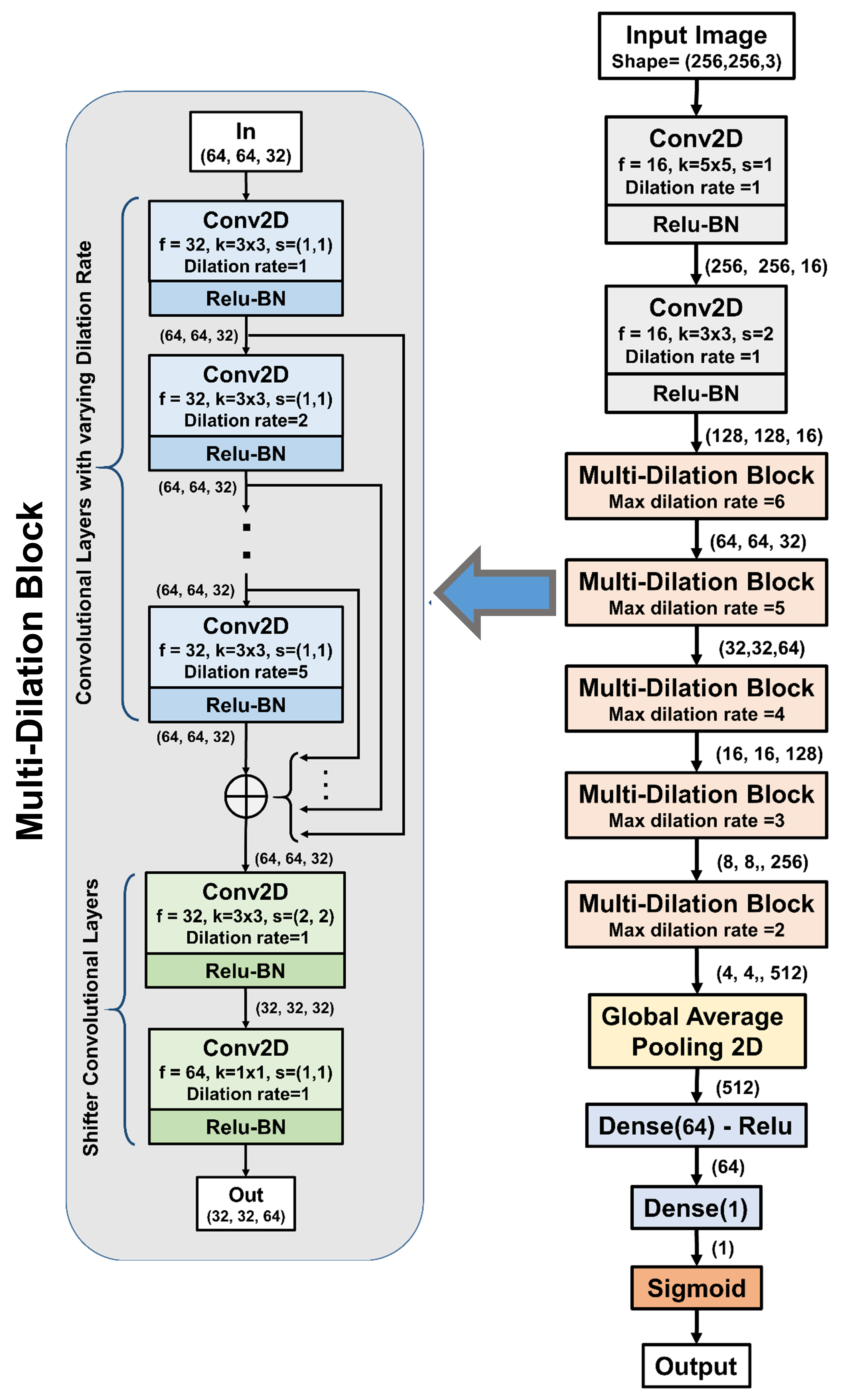}
\caption{\textbf{Proposed DilationNet architecture: A multi- dilation block is repeatedly used to extract features with varying dialtion rate.  This architecture is optimized for input images with shapes $(256,256,3)$. Here, `f' stands for number of filters, `k' for kernel size and `s' for the strides in the convolution operation.}}
\label{f3}
\end{figure}

The proposed DilationNet architecture is mainly introduced to efficiently optimize images with different resolutions. The detailed schematic of the proposed architecture is shown in Fig.~\ref{f3}. It is optimized for images with resolutions of $(256,256,3)$, while it can be optimized for other resolutions by changing the number of structural units according to the dimension of the transformed feature map.

In DilationNet, firstly, two layers of convolutions are used for initial processing of the images with larger kernels of $(5\times 5)$  and for reducing spatial dimension of the feature map for further processing with strided convolution, respectively. Later, series of multi-dilation blocks are introduced to process the feature map through a number of convolution operations. Each of such block generalizes the feature map by reducing spatial dimensions to broaden the observation level in later stages with an increase of the number of channels to introduce more filtering operations.

\begin{figure}[!t]
    \centering
    \subfloat[]{\includegraphics[scale=0.25]{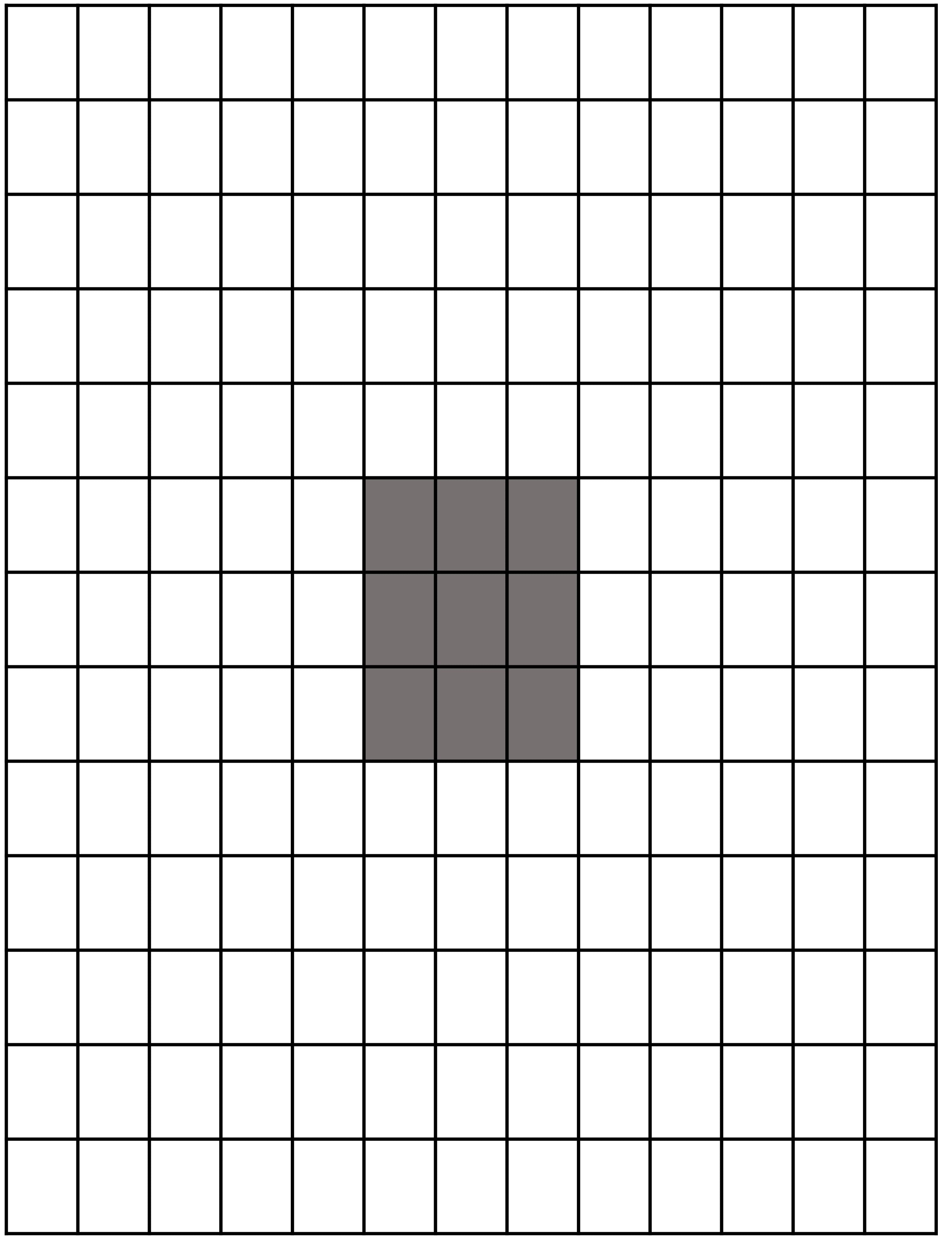}%
    \label{f2a}}
    \hspace{2mm}
    \subfloat[]{\includegraphics[scale=0.25]{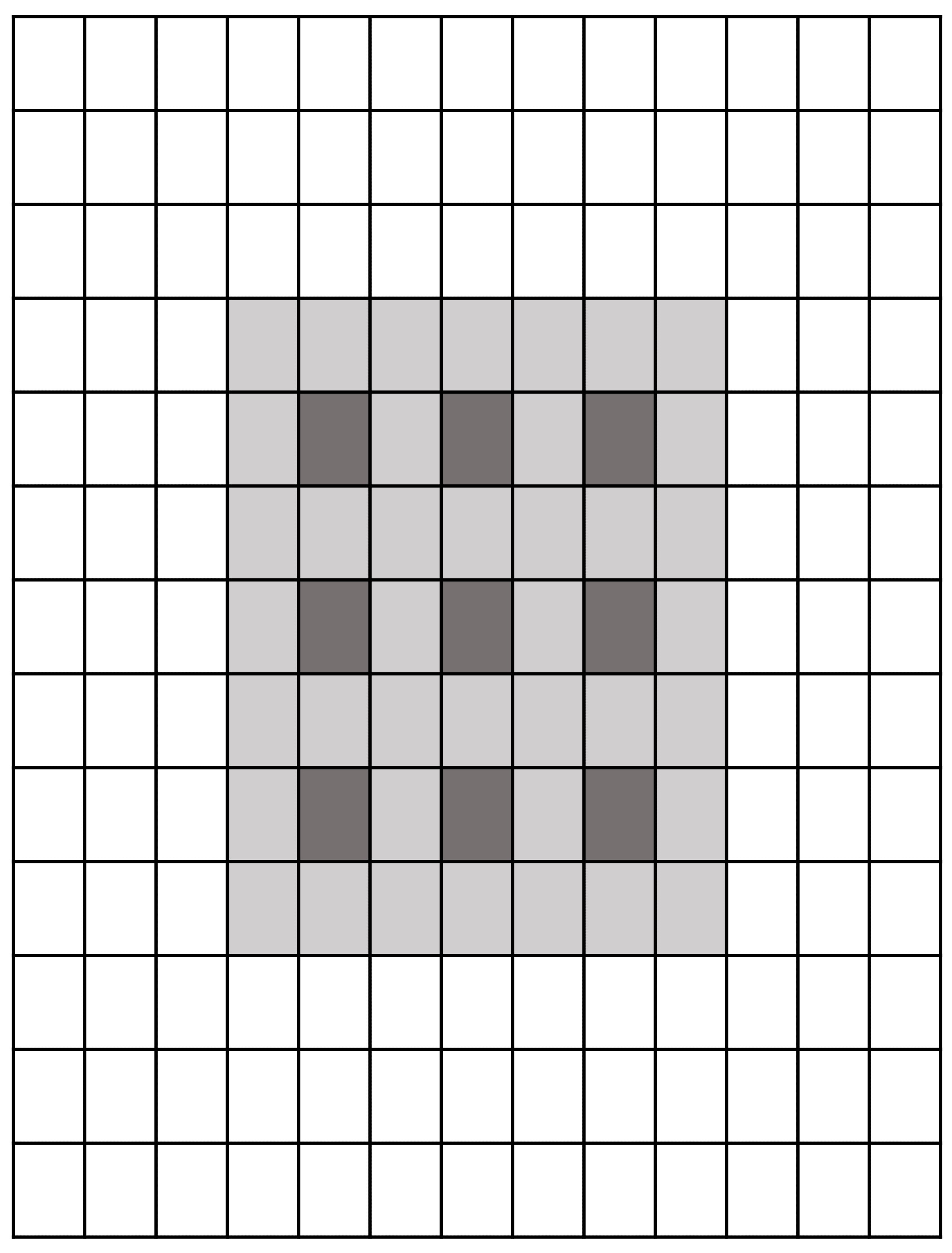}%
    \label{f2b}}
    \hspace{2mm}
    \subfloat[]{\includegraphics[scale=0.25]{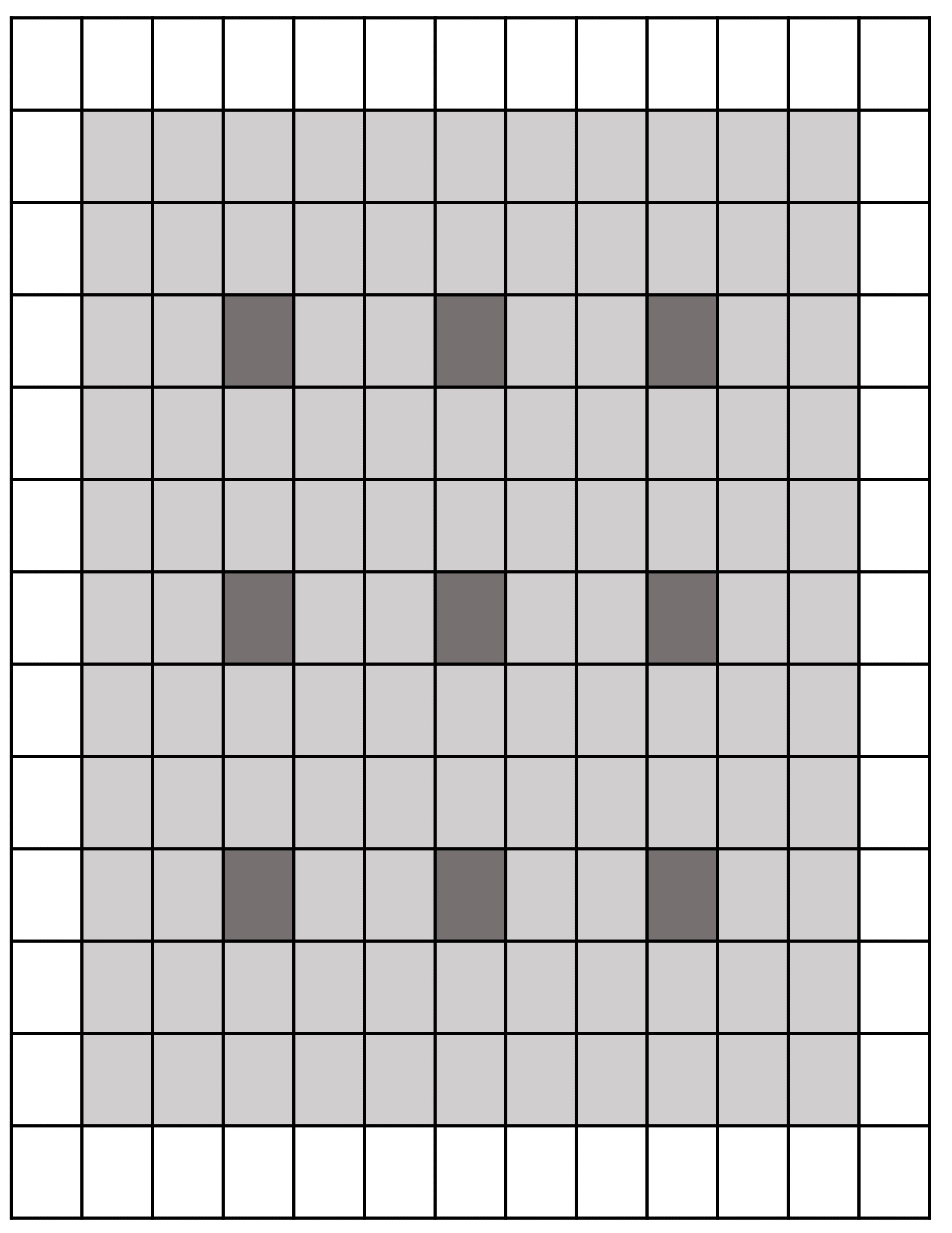}%
    \label{f2c}}
    \caption{\textbf{Convolutions with kernel size of $(3\times 3)$ with different dilation rate covering different receptive areas. (a) dilation rate = 1, (b) dilation rate=2, and (c) dilation rate=3.}} 
    \label{f2}
\end{figure}

In the multi-dilation block, series of convolutions operations are carried out while varying the dilation rate of the respective kernels. This type of dilated convolutions helps to extract information from a broader receptive area without increasing the size of the kernel and computation. In Fig.~\ref{f2}, such operation is presented schematically. It can be defined by.

\begin{equation}
    y(i,j) = \sum_p \sum_q x[i+ dp, j+dq] w[p,q]
\end{equation}

where $y$ is the output of convolution with dilation rate of $d$, $(i, j)$ represents the center of the convolution, $x$ represents the input feature map and $w$ is the filter/kernel.

Moreover, in the multi-dilation block output of each dilated convolution undergo further convolution operation with increased dilation rate. These operations can be represented by,
\begin{align}
    y^{k}(i,j) &= Relu \left(\sum_p \sum_q y^{k-1}[i+ kp, j+ kq] w[p,q] \right)\\
    &\forall{\ k \in \{1, 2, \dots, m\}} \nonumber
\end{align}

where $m$ is the maximum dilation rate and $y^{0}$ represents the input feature map of the multi-dilation block. As the feature map size reduces in the subsequent multi-dilation blocks, maximum dilation rate is reduced accordingly to adjust the maximum receptive area. Moreover, in between such dilated convolutions, rectified linear unit (Relu) is used as activation function to introduce non-linearity, which is given by,  

\begin{equation}
    Relu(x) =\left\{
    \begin{array}{cc}
        x, & x>0  \\
        0, & x<=0
    \end{array} \right.
\end{equation}

Additionally, batch normalization operation~\cite{BN} is also incorporated in between convolutional layers following the non-linear activation to accelearte the model convergence, which is given by, 
\begin{equation}
    BatchNormalization(\mathbf{X}) = \gamma \left( \frac{\mathbf{X}-E(\mathbf{X})}{\sqrt{Var(\mathbf{X})}} \right)   + \beta
\end{equation}

where $\gamma$ and $\beta$ are the two learnable hyper-parametrs for each layer. 

Afterward, all the outputs of convolutions with varying dilation rates are added together to produce $y_{sum}$ and thus,

\begin{equation}
    y_{sum} = Add(y_1, y_2, \dots,  y_m)
\end{equation}

These types of addition operations introduce flexibility in the feature extraction process to incorporate features from varying receptive fields. If any of the observed receptive fields provide more valuable information, it gets emphasized while other layers with modified receptive fields de-emphasize their outputs if they don't contribute valuable features in the training process. In such a way, each multi-dilation block explores a broader observation spectrum to effectively incorporate variegated features ranging from high frequency, localized features to low-frequency, generalized features.

Finally, two convolutions in series are carried out to reduce the spatial dimension and increase the channel of the feature map, respectively, for producing the output $(y_{out})$ from the multi-dilation block. Hence,

\begin{figure*}[!t]
\centering
\includegraphics[scale= 0.6]{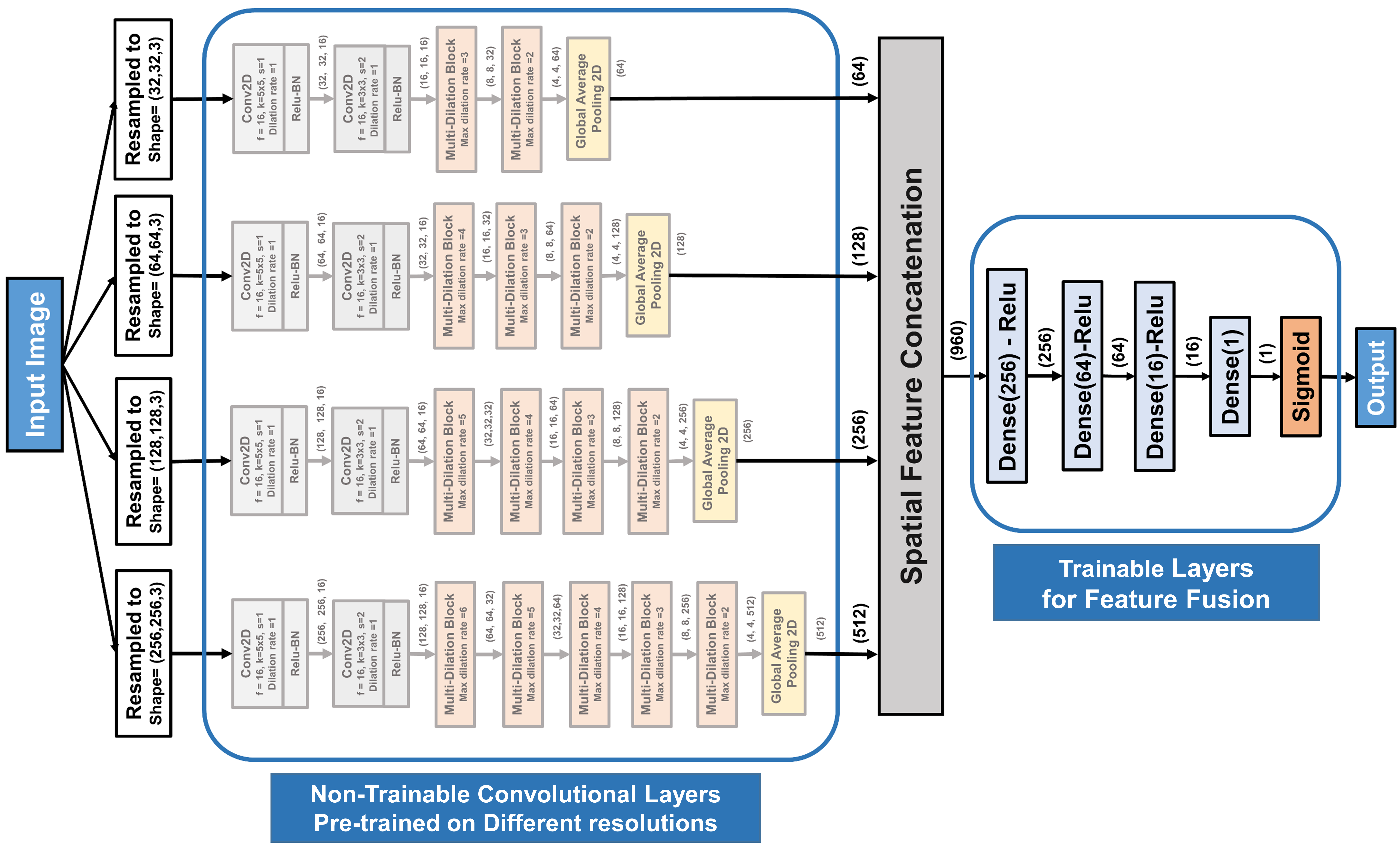}
\caption{\textbf{Proposed DeepFusionNet Architecture: Variegated feature spaces obtained from the stack of convolutional layers of different DilationNet architectures, initially optimized for various resolutions of images, are jointly optimized by training the stack of densely connected layers to make the effective fusion of features.}}
\label{f4}
\end{figure*}

\begin{equation}
    y_{out}= N_2(N_1(y_{sum}; \theta_1); \theta_2)
\end{equation}

where $N_1$ and $N_2$ represents these shifter convolutional layers to introduce dimension shift in the feature map and $\theta_1$ and $\theta_2$ are their respective parameters for optimization. 

After stacking series of such multi-dilation blocks, a global average pooling layer is used to extract global average of each individual channel of the feature map, which is given by,
\begin{align}
    GlobalAveragePooling(i) &= \frac{1}{P\times Q}\sum_{p=1}^{P} \sum_{q=1}^{Q} X(p,q,i)\\
    &\forall{\ i \in \{1, 2, \dots, N \}} \nonumber
\end{align}

where $(P,Q)$ represents the spatial dimension of $i_{th}$ channel of the feature map $X$ and $N$ is the total number of channels.

The total number of such multi-dilation blocks is changed with the variation in the resolution of input images along with the maximum dilation rate in each unit. This stack continues to rise until reaching the minimum spatial dimension of $(4,4)$ and the maximum dilation rate $m$ is also adjusted according to the spatial dimension. Hence, to operate with four different resolutions of images, different sets of convolutional layers will be employed. The output of the stack of convolutional layers, $f$, can be represented by,  

\begin{equation}
    f_r = F_r(X'_r);\ \forall{r \in\{32,64,128,256\}}
\end{equation}

where $F_r$ represents the stack of convolutional layers for resolution $r$ .

Afterwards, output of the convolutional operation, $f$, is being passed through series of densely connected layers with nonlinear activations in between to produce the final prediction, which is given by, 
\begin{align}
    y_{predicted, r} &=  Sigmoid(Relu(\mathbf{W_1}*f_r + \mathbf{b_1})* \mathbf{W_2} + \mathbf{b_2}) \nonumber\\
                &= D_r(f_r) ;\ \forall{r \in\{32,64,128,256\}}
\end{align}

where $\mathbf{W}$ and $\mathbf{b}$ represents the weight vector and the bias vector, respectively, of the densely connected layers, and $D_r$ represents the stack of densely connected layers for resolution $r$. Finally, sigmoid activation is used to provide the final prediction, which is given by,
\begin{equation}
    Sigmoid(x) = \frac{1}{1 + e^{-x}}
\end{equation}

\subsection{Multi Resolution Feature Fusion with Proposed DeepFusionNet:}
In the initial individual training stage, all the developed forms of various DilationNet architectures are trained and optimized to efficiently extract representational features to diagnose malaria. In this stage, a feature fusion scheme will be introduced with the proposed DeepfusionNet architecture (Fig.~\ref{f4}) for joint optimization of the diverse feature space with features from different levels of observations using all the trained DilationNets.

As all the stacks of convolutional layers in different DilationNets are optimized to extract spatial features starting from different resolutions of images, output of all these layers will provide the most generalized and optimized representation of features from respective levels of observations. Hence, for each input image, varieties of feature spaces will be obtained from output of different stacks of convolutional layers of various DilationNets. These variegated features are concatenated to produce the fusion feature vector, $F_{fusion}$, which is given by,  

\begin{equation}
    F_{fusion} = Concatenate(f_{32}, f_{64}, f_{128}, f_{256})
\end{equation}

Afterwards, this fusion feature vector undergoes through joint optimization by a series of densely connected layers with 
nonlinear activation functions in between, which can be represented by,
\begin{align}
    d_i &= \sigma(\mathbf{W_i}* d_{i-1} + \mathbf{b_i})\\
    &\forall{i \in \{1,2,3,4\}} \nonumber
\end{align}

where $d_i$ is the output of the $i_{th}$ densely connected layer with $W_i$ weight vector and $b_i$ bias vector, $\sigma$ represents the activation function and $d_0= F_{fusion}$. In total, four densely connected layers are stacked in series for converging the fusion vector toward the final prediction. 

Finally, the prediction of this DeepFusionNet architecture, $y_{predicted, fusion}$, is obtained by feeding the output of the final densely connected layer $d_4$ into the sigmoid function, and thus, 

\begin{equation}
    y_{predicted, fusion} = Sigmoid(d_4)
\end{equation}
    
\subsection{Network Training and Optimization}
All the individual DilationNets along with the DeepFusionNet are trained by using back propagation algorithm with $L2$ regularization to minimize the cross entropy loss function $(\mathscr{L})$, which is given by,

\begin{equation}
    \begin{split}
    & \mathscr{L}(W)= -\frac{1}{n} \biggl[ \sum_{i=1}^{n} \sum_{k=1}^{K} y_k^{(i)} \ln{(h_w(x^{(i)}))}_k +\\ 
    &+ (1-y_k^{(i)}) \ln{(1-(h_w(x^{(i)})_k)} \biggr] + \frac{\lambda}{2n} \sum_w w^2    
    \end{split}
\end{equation}

where  $W$ denotes the weight vector, $y_k^{(i)}$ is the actual level, $h_w(x^{(i)})_k$ is the predicted output at $k_{th}$ node of output for $i_{th}$ input and $\lambda$ is the regularization parameter to reduce overfitting.

In the training phase, the weight $w_j$ of the $j_{th}$ layer at time t is optimized using Adam optimizer~\cite{adam}  with learning rate of $\eta$, which is given by,

\begin{equation}
    w_{t+1}^j = w_t^j - \eta \frac{v_t}{\sqrt{s_t+\epsilon}} \times g_t
\end{equation}
\begin{align}
    \text{where, }&v_t = \beta_1\times v_{t-1}-(1-\beta_1)\times g_t \\
    &s_t = \beta_2\times s_{t-1}-(1-\beta_2)\times g_t^2
\end{align}

Here, $g_t$ is gradient at time $t$, $v_t$ and $s_t$ are the exponential average of gradients and squares of gradients along $w_j$, respectively, at time t, and $\beta_1$ and $\beta_2$ are the two hyperparameters that are chosen empirically to be $0.9$ and $0.99$, respectively, for faster convergence.

\begin{table*}[t]
\centering
\caption{Performance of the DeepFusionNet Architecture with Fusion of Features from Different Combination of DilationNet Architectures Operating with Various Resolutions of Images}
\label{t1}
    \begin{tabu}{|[1.25pt]c|[1pt]c|[1pt]c|[1pt]c|[1pt]c|[1pt]c|[1.25pt]}
            \tabucline[1.25pt]{-}
            \textbf{Network Combination} & \textbf{Accuracy (\%)} & \textbf{Sensitivity (\%)} & \textbf{Specificity (\%)} & \textbf{Kappa Score (\%)} & \textbf{AUC Score (\%)} \\
            \tabucline[1.25pt]{-}
            \textbf{A+B}                 & 97.8                   & 97.7                      & 97.2                      & 95.9                      & 99.5                    \\
            \hline
            \textbf{A+C}                 & 98.1                   & 97.9                      & 97.7                      & 96.8                      & 99.6                    \\
            \hline
            \textbf{A+D}                 & 98.3                   & 98.3                      & 98                        & 97.2                      & 99.7                    \\
            \hline
            \textbf{B+C}                 & 98.5                   & 98.1                      & 97.5                      & 97.1                      & 99.7                    \\
            \hline
            \textbf{B+D}                 & 98.6                   & 98.4                      & 97.9                     & 97.3                      & 99.7                    \\
            \hline
            \textbf{C+D}                 & 98.7                   & 98.5                      & 98.2                      & 97.8                      & 99.8                    \\
            \hline
            \textbf{A+B+C}               & 98.7                   & 98.4                      & 98.1                      & 97.7                      & 99.7                    \\
            \hline
            \textbf{A+B+D}               & 98.9                   & 98.8                      & 98.3                      & 98.2                      & 99.8                    \\
            \hline
            \textbf{A+C+D}               & 99.1                   & 99                        & 99.2                      & 98.4                      & 99.8                    \\
            \hline
            \textbf{B+C+D}               & 99.3                   & 99.2                      & 99                  & 98.7                      & 99.8                    \\
            \hline
            \textbf{A+B+C+D}             & 99.5                   & 99.6                      & 99.5                      & 99.1                      & 99.9                    \\              
            \tabucline[1.25pt]{-}
        \end{tabu}     
\end{table*}

\section{Results and Discussions}
In this section, experimental results are analyzed to verify the proposed scheme. A publicly available dataset is used for extensive experimentation. performance obtained from different training stages with the various resolution of images along with the feature fusion schemes are comparatively analyzed in detail. Finally, to confirm the effectiveness of the proposed scheme, the performance of other state-of-the-art methods are compared. 
\begin{figure}[!t]
\centering
\includegraphics[scale= 0.39]{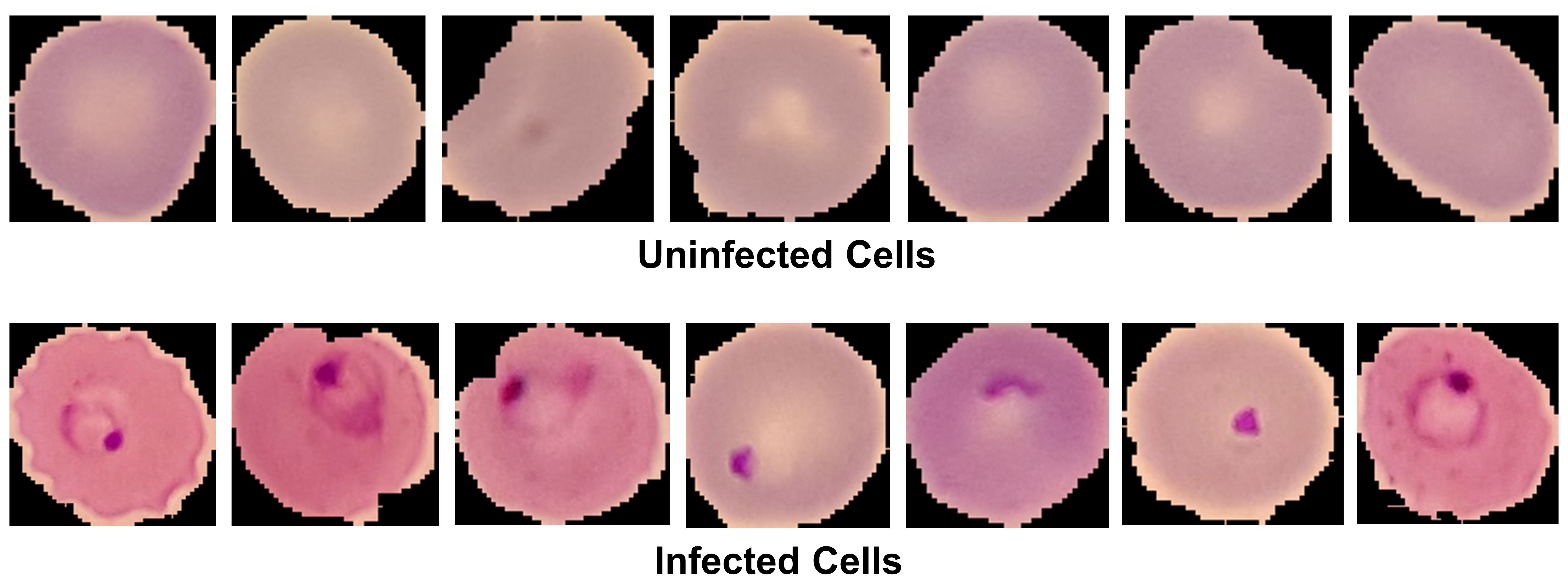}
\caption{\textbf{Sample pictures from the database: Thin blood smear images of some of the uninfected cells and some of the infected cells are shown.}}
\label{f5}
\end{figure}

\begin{figure}[!t]
\centering
\includegraphics[scale= 0.38]{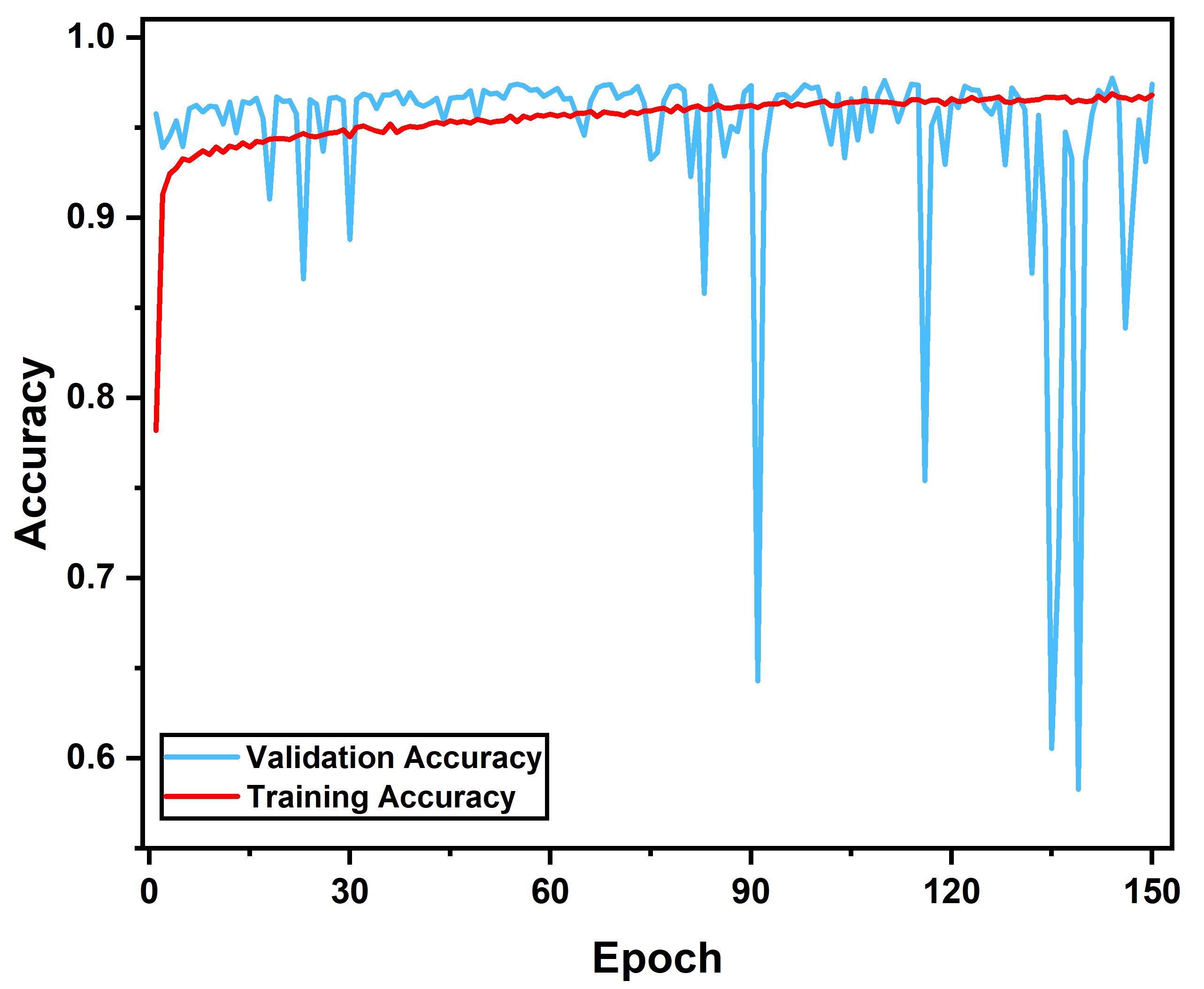}
\caption{\textbf{Training/Validation accuracy of DilationNet-C over epochs for resolution $(128\times128)$: The convergence of the DilationNet architecture is very fast. Validation accuracy achieves considerable value in a few numbers of training epochs.}}
\label{f6}
\end{figure}
\subsection{Database Description}
The public database provided by the National Institute of Health (NIH) is used in this work~\cite{nih}. It consists of segmented cells of thin blood smear slide images collected from several healthy patients and Plasmodium falciparum-infected patients at Chittagong Medical College Hospital, Bangladesh. It contains $13,779$ parasitized and $13,779$ uninfected cell images manually annotated by experts. some of the sample images of both the infected cells and uninfected cells are shown in Fig.~\ref{f5}. $80\%$ of the total images are randomly selected for training the network while remaining $20\%$ are used as the test set for evaluating performance. To make the training more robust, a number of traditional augmentation techniques are applied to the training data leaving the test set unaltered for proper validation. A random combination of all these methods is utilized for introducing some realistic random variations on the training data.  

\subsection{Experimental Setup}
Intel\textregistered\  Xeon\textregistered \ $D-1653N$ CPU @$2.80$GHz with $12$M Cache and $8$ cores is used for experimentation. For hardware acceleration, $1$ x NVIDIA Titan RTX GPU having with $4608$ CUDA cores running $1770$ MHz with $24$ GB GDDR6 memory is deployed.
Numerous traditional metrics of binary classification are used for evaluating the performance of the proposed architectures such as accuracy, sensitivity, specificity, area under curve (AUC) score, and Cohen-Kappa score. These are defined by,
\begin{align}
    &\text{Accuracy} = \frac{TP+FP}{TP+TN+FP+FN} \\
    &\text{Sensitivity}= \frac{TP}{TP+FN} \\
    &\text{Specificity} = \frac{TN}{TN+FP} \\
    &\text{Kappa Score} = \frac{P_o-P_e}{1-P_e}
\end{align}

\begin{figure}[!t]
\centering
\includegraphics[scale= 0.37]{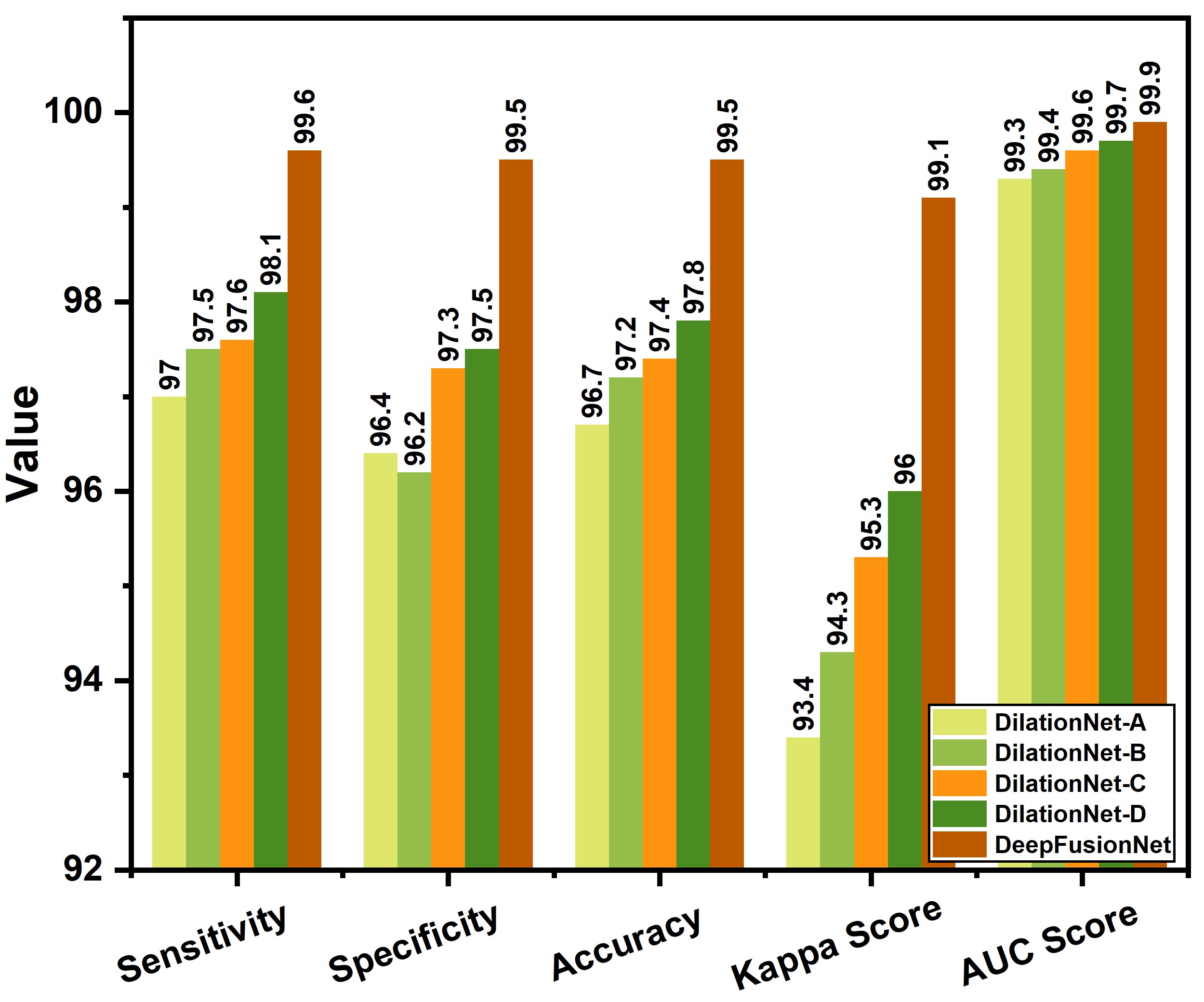}
\caption{\textbf{Comparison of performance of different forms Dilationnet architectures with the DeepFusionNet architecture: Performance of DilationNet gets better with increasing resolutions of images. Considerable increase of performance is achieved by making the feature fusion with DeepFusionNet.}}
\label{f7}
\end{figure}

\begin{figure}[!t]
\centering
\includegraphics[scale= 0.34]{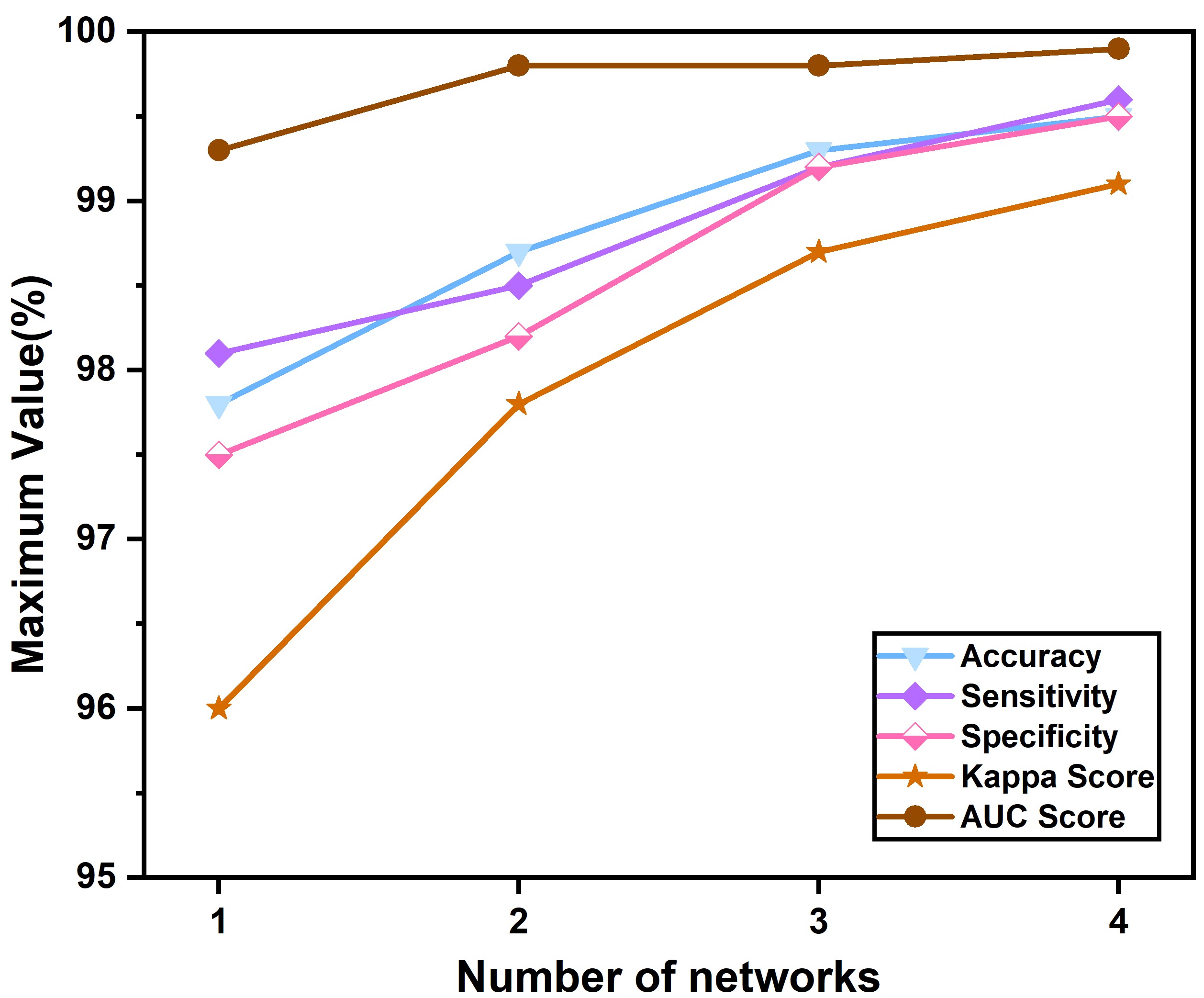}
\caption{\textbf{Change of maximum obtained value in different evaluation metrics with increasing number of networks in DeepFusionNet: Performance of the DeepFusionNet improves with increasing number of networks for increasing the variations in the feature fusion scheme.}}
\label{f8}
\end{figure}

\begin{figure}[!t]
\centering
\includegraphics[scale= 0.34]{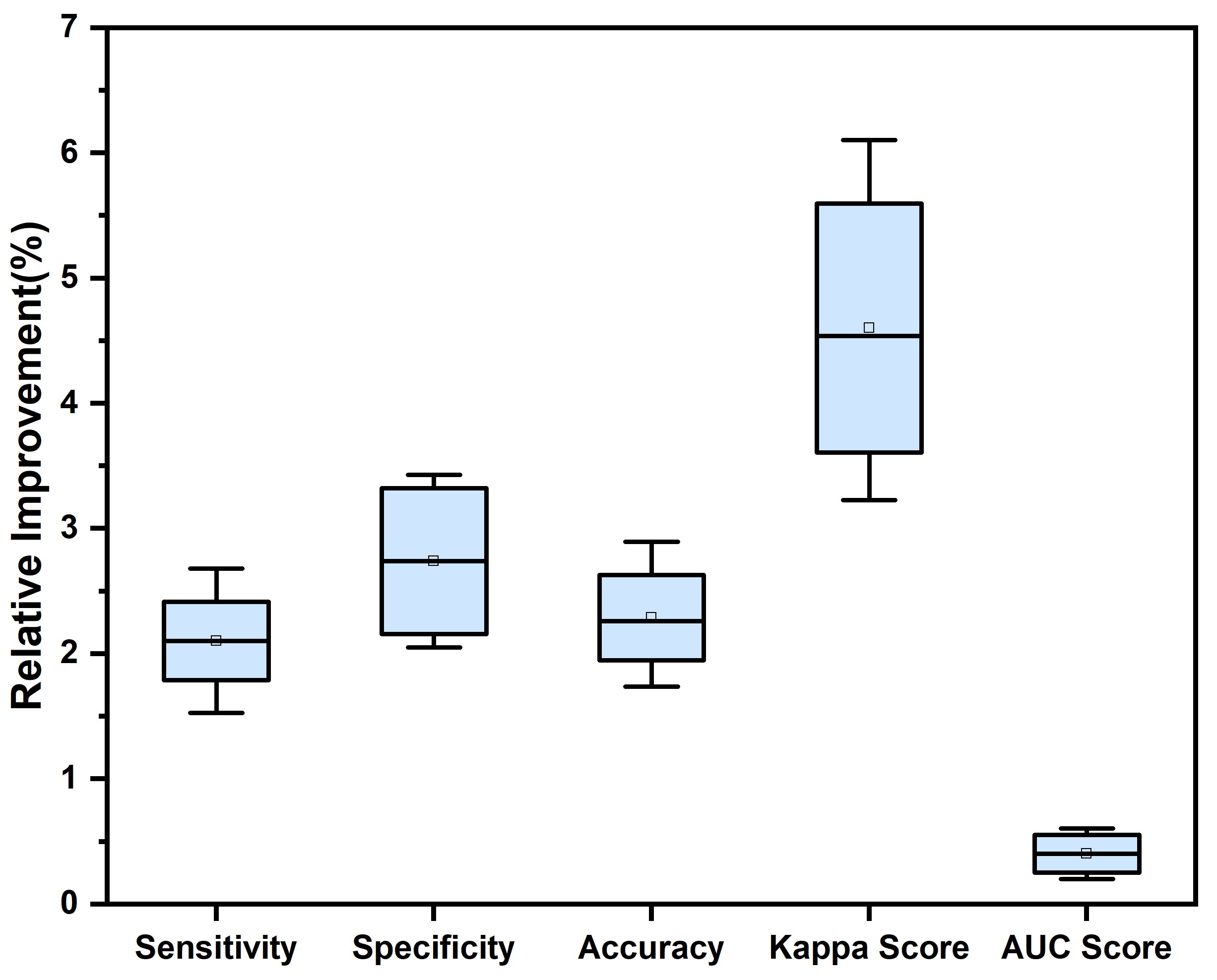}
\caption{\textbf{Variations of relative improvements in DeepFusionNet with respect to the individual networks operating with different resolutions: All the metrics provide improvement in the fusion scheme from the individual networks. Kappa score provides the largest variance in relative improvements for the steepest rise.}}
\label{f9}
\end{figure}

\begin{table*}[htbp]
\centering
\caption{Performance of the Proposed Feature Fusion Scheme for Training with Different Amounts of Available Data}
\label{t2}
\begin{tabu}{|[1.25pt]c|[1pt]c|[1pt]c|[1pt]c|[1pt]c|[1pt]c|[1.25pt]}
\tabucline[1.25pt]{-}
\textbf{Percentage of the Dataset} & \textbf{Accuracy (\%)} & \textbf{Sensitivity (\%)} & \textbf{Specificity (\%)} & \textbf{Kappa Score (\%)} & \textbf{AUC Score (\%)} \\
\tabucline[1.25pt]{-}
\textbf{10\%}                      & 95.6                   & 96.1                      & 94.2                      & 91.4                      & 97.3                    \\ \hline
\textbf{20\%}                      & 96.3                   & 96.9                      & 95.1                      & 93.1                      & 97.7                    \\ \hline
\textbf{30\%}                      & 97.1                   & 97.4                      & 96.8                      & 94.9                      & 98.2                    \\ \hline
\textbf{40\%}                      & 98.2                   & 97.6                      & 97.1                      & 96.1                      & 98.4                    \\ \hline
\textbf{50\%}                      & 98.4                   & 97.9                      & 97.5                      & 96.8                      & 98.7                    \\ \hline
\textbf{60\%}                      & 98.8                   & 98.1                      & 97.9                      & 97.4                      & 98.8                    \\ \hline
\textbf{70\%}                      & 99.1                   & 98.7                      & 98.3                      & 98.1                      & 98.9                    \\ \hline
\textbf{80\%}                      & 99.3                   & 99.1                      & 98.9                      & 98.7                      & 99.0                    \\ \hline
\textbf{90\%}                      & 99.5                   & 99.4                      & 99.1                      & 98.9                      & 99.1                    \\ \hline
\textbf{100\%}                     & 99.5                   & 99.6                      & 99.5                      & 99.1                      & 99.1                    \\ 
\tabucline[1.25pt]{-}
\end{tabu}
\end{table*}

\subsection{Performance Analysis}
Firstly, different resolutions of blood smear images are used for optimizing the different forms of DilationNet architecture. These variants of DilationNet are used to explore the data from different levels to extract effective features for providing better performance. In the proposed study, four different forms of DilationNet are produced to operate with images of four different resolutions. Each of the proposed DilationNet architectures is designed for further exploration of the feature space with introduced variations in the receptive fields by utilizing varying dilation rates in the convolutions. This architectural scheme provides the opportunity to investigate images of any specific resolution with diversified perspectives that makes the feature extraction process extremely efficient. In Fig.~\ref{f6}, both training accuracy and validation accuracy are shown for the DilationNet-C operating with resolutions of $(128\times128)$. It should be noticed that the validation accuracy has been reached to a considerable limit within a few epochs. It validates the effectiveness of the proposed DilationNet architecture that can be optimized faster for its effective feature extraction scheme by proper utilization of numerous convolutions with varying dilation rates.

Afterward, all the forms of DilationNets are optimized with the DeepFusionNet by introducing a feature fusion scheme to obtain the global maxima of the evaluation metrics. Different combinations of the DilationNet architectures are experimented in the fusion scheme using DeepFusionNet. Each of the DilationNet architectures adds some valuable features after in detail exploration of the input image of a specific resolution. It is expected that by fusing the extracted features from all these networks will increase the feature exploration as well as feature extraction capacity of the network. In Table~\ref{t1}, the performance of the fusion scheme with different combinations of DilationNets is presented. It can be implied that with the addition of multiple networks in the fusion scheme, more diversity is integrated into the fusion schemes that provide the continuing rise of performance in all evaluation metrics.  As some of the combinations of networks provide overlapping features with fewer variations that result in comparable smaller values in different metrics. In Fig.~\ref{f8}, the effect of the integration of a different number of networks in the fusion scheme is plotted. Maximum values achieved for combinations of the different number of networks are considered. It should be noted that with an increasing number of networks the performance continues to get better. However, as more networks are integrated, the rate of increase in performance slows down due to the incorporation of smaller diversity when features from a large spectrum are already explored with other networks.

After the integration of all the four networks in the DeepFusionNet, the best performance is achieved. In Fig.~\ref{f7}, the performances of all the individual DilationNets are shown along with the DeepFusionNet. It should be easily noticed that the fusion scheme provides a significant increase in performance compared to all the individual networks. In Fig.~\ref{f9}, relative improvements of performance in the DeepfusionNet are shown with respect to the individual networks. With increased diversity in the feature space, a sharp increase is noticeable in most of the evaluation metrics. However, the kappa score provides the steepest increase with the fusion scheme while the AUC score is increased by a comparably smaller margin due to its higher obtained value with the individual networks. 

\begin{figure}[!t]
\centering
\includegraphics[scale= 0.36]{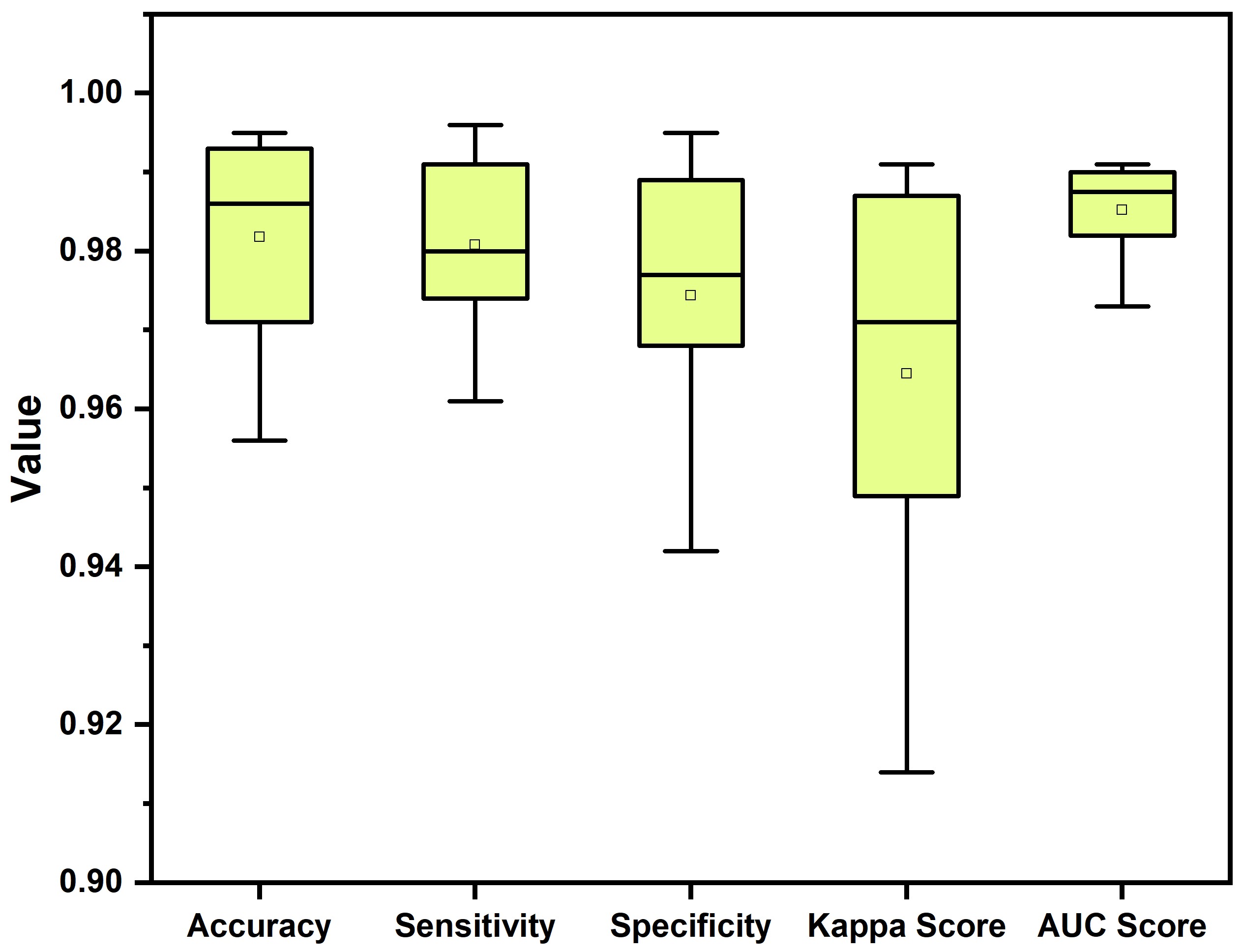}
\caption{\textbf{Variations of performance in different metrics for training with increasing amount of available data: Considerable performance is achieved even with $10\%$ of the available training data. With increasing data, performanmce gets better in all the metrics.}}
\label{f10}
\end{figure}

\begin{table*}[t]
\centering
\caption{Comparison of Performance of the Proposed Deep Feature Fusion Scheme with Other State-of-the-Art Methods}
\label{t3}
\begin{tabu}{|[1.25pt]c|[1pt]c|[1pt]c|[1pt]c|[1pt]c|[1pt]c|[1.25pt]}
\tabucline[1.25pt]{-}
\textbf{Method}        &        \textbf{Accuracy (\%)} & \textbf{Sensitivity (\%)} & \textbf{Specificity (\%)} & \textbf{Kappa Score (\%)} & \textbf{AUC Score (\%)} \\
\tabucline[1.25pt]{-}
\textbf{ResNet-50~\cite{res}}              & 95.7                   & 94.5                      & 96.9                      & 93.2                      & 99.0                    \\
\hline
\textbf{DenseNet-121~\cite{dense}}           & 93.1                   & 94.2                      & 92.6                      & 91.6                      & 97.6                    \\
\hline
\textbf{VGG-16~\cite{vgg}}                 & 94.5                   & 93.9                      & 95.1                      & 92.5                      & 98.1                    \\
\hline
\textbf{Rajarman~\emph{et al.}~\cite{nih}}        & 95.9                   & 94.7                      & 97.2                      & -                         & 99.1                    \\
\hline
\textbf{Gopakumar~\emph{et al.}~\cite{c4}}       & 97.7                   & 97.1                      & 98.5                      & -                         & -                       \\
\hline
\textbf{Bibin~\emph{et al.}~\cite{belief}}           & 96.3                   & 97.6                      & 95.9                      & -                         & -                       \\
\hline
\textbf{Liang~\emph{et al.}~\cite{c3}}           & 97.3                   & 96.9                      & 97.7                      & -                         & -                       \\
\hline
\textbf{Das~\emph{et al.}~\cite{c2}}             & 84.0                   & 98.1                      & 68.9                      & -                         & -                       \\
\hline
\textbf{Proposed DeepFusionNet} & \textbf{99.5}        & \textbf{99.6}           & \textbf{99.5}           & \textbf{99.1}           & \textbf{99.9}         \\
\tabucline[1.25pt]{-}
\end{tabu}
\end{table*}

An additional experiment is designed to explore the robustness of the proposed scheme with a smaller amount of available data. Performance of the proposed fusion scheme with different percentage of the available data is provided in Table~\ref{t2}. It should be noted that the proposed scheme achieved a considerable performance of $95.6\%$ accuracy even with $10\%$ of the available training images. However, with the integration of more amount of data, the network continues to perform better than before. In Fig.~\ref{f10}, variations of the achieved performance in different metrics are shown for introduced variations in the available training data. These metrics approach closer to the perfect value as more amount of data are utilized. Hence, it can be said that the proposed scheme is capable of handling a smaller amount of data effectively as well as contains the ability to achieve finer performance with more amount of data.

Finally, the proposed scheme is compared with some of the state-of-the-art approaches along with some traditional, efficient neural networks that are heavily used in numerous computer vision applications. it should be noted that most of the other approaches used heavy pre-processing schemes to achieve higher accuracy while the proposed scheme used minimal amount of pre-processing to make the testing phase faster and simpler. It can be clearly noticed that the proposed scheme outperforms all other approaches by a significant margin in all the evaluation metrics that further validate the effectiveness of the proposed scheme for precise diagnosis of malaria.

\section{Conclusion}
Multi-resolution feature fusion scheme is proposed which aims to increase the diversity of the extracted features by incorporating images with different resolutions along with the multi-receptive feature extraction process with varying dilation rates in the convolution. The designed DilationNet architecture is highly scalable with adaptive design specifications to adjust images of different resolutions. It is shown that this network is highly efficient that achieves considerable accuracy in fewer epochs. Afterward, the introduced feature fusion scheme with the proposed DeepFusionNet architecture provides the opportunity for further exploration of the optimized feature spaces of different forms of DilationNet. It is observed that with the increasing number of networks in the fusion scheme, the performance continues to improve for the integration of more diversified features extracted from different observation windows.  In addition, it is also found that the proposed scheme can achieve considerable performance with a very small amount of training data as well as can be optimized for improved performance with the integration of more data. Through extensive experimental analysis on a publicly available dataset, a significant increase in performance is achieved by the proposed scheme compared to other state-of-the-art approaches. This scheme opens up the immense opportunity for exploring the diversified feature spaces generated by numerous individually trained networks operating in various levels of representation of the data to converge into the most optimized representations that result in precise diagnosis. Therefore, the proposed method can be applied to efficiently diagnose malaria from thin blood smear images as well as can be optimized for other image-based disease diagnosis applications.

\bibliographystyle{IEEEtran}
\bibliography{references}

\end{document}